\documentclass{article}
\usepackage{ijcai21}

\usepackage{times}

\usepackage{soul}
\usepackage{url}
\usepackage[hidelinks]{hyperref}
\usepackage[utf8]{inputenc}
\usepackage[small]{caption}
\usepackage{graphicx}
\usepackage{amsmath}
\usepackage{booktabs}
\usepackage{multirow}
\usepackage{multicol}
\usepackage{algorithm}
\usepackage{algorithmic}
\usepackage{subcaption}
\usepackage{enumitem}
\usepackage{makecell}
\usepackage{amssymb}
\usepackage{xcolor}

\DeclareMathOperator*{\argmin}{arg\,min}

\title{Temporal Knowledge Graph Completion: A Survey}

\author{Borui Cai$^1$\and
Yong Xiang$^1$\and
Longxiang Gao$^{1}$\and
He Zhang$^2$\and
Yunfeng Li$^2$\And
Jianxin Li$^{1}$

\affiliations
$^1$School of Information Technology, Deakin University\\
$^2$CNPIEC KEXIN, Ltd
\emails
\{b.cai,yong.xiang,longxiang.gao,jianxin.li\}@deakin.edu.au,
\{zhanghe,liyunfeng\}@cnpiec.com.cn}

\begin{document}

\maketitle

\begin{abstract}
Knowledge graph completion (KGC) can predict missing links and is crucial for real-world knowledge graphs, which widely suffer from incompleteness. 
KGC methods assume a knowledge graph is static, but that may lead to inaccurate prediction results because many facts in the knowledge graphs change over time. 
Recently, emerging methods have shown improved predictive results by further incorporating the timestamps of facts; namely, temporal knowledge graph completion (TKGC). 
With this temporal information, TKGC methods can learn the dynamic evolution of the knowledge graph that KGC methods fail to capture.
In this paper, for the first time, we summarize the recent advances in TKGC research. 
First, we detail the background of TKGC, including the problem definition, benchmark datasets, and evaluation metrics. 
Then, we summarize existing TKGC methods based on how timestamps of facts are used to capture the temporal dynamics. 
Finally, we conclude the paper and present future research directions of TKGC.

\end{abstract}

\section{Introduction}
Knowledge graphs are multi-relational graphs that represent a wide range of real-world events as structured facts. A fact is composited of two entities (as nodes) and the relation that connects them (as the edge). The abundant information carried by knowledge graphs have made them favorable for various applications, e.g., content-based recommender system \cite{recommender}, natural language question-answering \cite{question}, and text-centric information retrieval \cite{ir}. Unfortunately, despite their large scales, existing knowledge graphs (of both academic and industry) widely suffer from incompleteness. For example, in Freebase, one of the prestigious public knowledge databases, more than 70\% of $person$ entities have unknown $place\ of\ birth$ \cite{70}. 
This data scarcity issue greatly limits the effectiveness of knowledge graphs for downstream applications. 

Knowledge graph completion (KGC) \cite{kgc} aims at automatically inferring missing facts for a knowledge graph by learning from existing facts (also known as link prediction). 
Benefiting from the blooming of machine learning and deep neural networks, many KGC methods perform effective link prediction through knowledge graph embedding \cite{kgcsurvey}. That is, learning low-dimensional representations for entities and relations with factual score functions, which measure the correctness of a fact. 
Despite their successes, one major limitation of KGC methods is that they can hardly learn the temporal dynamics of facts since they assume facts are static; however, many facts change over time. For example, the fact $\{DonaldTrump, presidentOf, USA\}$ is only true from 2017 to 2021, ignoring such temporal information may lead to ambiguity and misunderstanding. Not to mention that the temporal dynamics of facts also carry essential causal patterns that can assist the link prediction.

Recently, temporal knowledge graph completion (TKGC) methods \cite{ttranse} start emerging, aiming at more accurate link prediction by addressing the above limitation of KGC methods. That is, in addition to the facts, TKGC methods further incorporate timestamps of facts into the learning process. TKGC methods widely have shown improved link prediction accuracy than KGC methods on multiple knowledge graph datasets. Specifically, the key challenge of TKGC is how to effectively integrate timestamps into the model, so that the temporal dynamics of entities, relations, and the underlying graph can be properly captured and used for the link prediction. Although TKGC is increasingly attractive to the research community, so far we have not found a published survey for TKGC. In this paper, we fill this gap by summarizing for the first time the progress of current TKGC research. The contributions of this paper are summarized as follows: 1) We propose a taxonomy of existing TKGC methods based on how timestamps of facts are integrated for the link prediction. 2) We provide a detailed analysis of existing TKGC methods, and summarize common benchmark datasets and the general evaluation protocol. 3) We discuss the limitations of existing TKGC methods and provide several future research directions for TKGC.

The rest of this paper is organized as follows. Section 2 provides the background of TKGC, including the problem definition, the available benchmark datasets, and the general evaluation protocol. Section 3 analyzes existing TKGC methods. Finally, Section 4 draws the conclusion and discusses future research directions of TKGC.

\section{Background}
In this section, we first define the problem of TKGC, and then introduce available benchmark datasets and the general evaluation protocol.

\subsubsection{Problem Definition}
A knowledge graph is a directed multi-relational graph that contains structured facts. A fact is composited of two entities, a relation that connects the entities, and a timestamp. A timestamp normally is a time point (e.g., $2010$) or a time interval (e.g., $2010-2013$). For convenience, many TKGC methods split time intervals into discrete time points \cite{cygnet}. Unless otherwise specified, we regard timestamps as time points in the rest content.

We denote a knowledge graph that contains timestamps as $\mathcal{G}=(\mathcal{E},\mathcal{R},\mathcal{T},\mathcal{F})$, where $\mathcal{E}$, $\mathcal{R}$ and $\mathcal{T}$ are the sets of entities, relations, and timestamps, respectively.
$\mathcal{F}\in \mathcal{E}\times\mathcal{R}\times\mathcal{E}\times\mathcal{T}$ is the set of all possible facts. 
A fact is denoted $s=\{h, r, t, \tau\}$, where $h$, $r$, $t$ and $\tau$ are the head entity, the relation, the tail entity, and the timestamp, respectively.
The collection of facts in a knowledge graph is denoted as $\mathcal{D}\in \mathcal{F}$. We use $e$ to denote a learned representation. A factual score function, $q(s)$, is used to measure the correctness of a fact for training.
Negative sampling \cite{negativesample1} creates negative samples $s'$ by randomly corrupting true facts, which are widely used by TKGC methods to improve the expressiveness of learned representations. A loss function aims at collectively minimizing $q(s)$ and maximizing $q(s')$ for all facts and their negative samples. We summarize three different losses commonly employed by TKGC methods. 

The margin ranking loss \cite{transe} is adopted to ensure large score margins between facts and their corresponding negative samples, which is defined as follows:
\begin{equation}
\ell=\sum_{s\in \mathcal{D}}\Big{[}q(s)-\frac{1}{|\mathcal{D}_{s}^{-}|}\sum_{s'\in\mathcal{D}_{s}^{-}}q(s')+\gamma\Big{]}_{+},
\label{eq:margin}
\end{equation}
where $[x]_{+}=max(x,0)$ and $\gamma$ is a hyper-parameter that regulates how large the score margin is expected. The cross entropy loss \cite{cluster} also aims at obtaining a large separation gap between facts and negative samples, but it does not enforce a fixed score margin for all facts:
\begin{equation}
\ell=\sum_{{s\in \mathcal{D}}}\frac{exp(q(s))}{\sum_{s'\in \mathcal{D}_{s}^{-}}exp(q(s'))}.
\label{eq:cross}
\end{equation}
The binary cross entropy loss \cite{context} emphasises the score of individual facts and negative samples as follows:
\begin{equation}
\ell=\sum_{x\in \mathcal{D}\cup \mathcal{D}_{s}^{-}}yq(x)+(1-y)q(x),
\label{eq:binarycross}
\end{equation}
where $y=1$ if $x\in \mathcal{D}$ and $y=0$ otherwise. The convenient calculation makes binary cross-entropy loss favorable for neural network-based TKGC methods.

\subsubsection{Benchmark Datasets}
We summarize benchmark datasets widely used for the evaluation of TKGC methods in Table \ref{tab:dt}. The ICEWS and GDELT datasets contain events with time points, which are respectively extracted from the Integrated Crisis Early Warning System repository \cite{icews} and the Global Database of Events, Language, and Tone \cite{GDELT}. 
WIKIDATA \cite{ttranse} contains events extracted from the Wikidata knowledge base, with timestamps as time intervals like ``occursSince 2013". YAGO15K \cite{ta-transe} augmentes events of FB15K \cite{transe} with time intervals similar to WIKIDATA; but is more challenging since many facts have no timestamp.
Time-aware filtering \cite{tango} is normally adopted to ensure reasonable evaluation. That means entities that cause ambiguity are removed from the candidate list for a query.
The performance of existing models on GDELT dataset is the poorest among these datasets, though it contains the largest number of facts and the smallest entity and relation sets. The reason partly is because GDELT contains many abstract/concept entities (e.g., GOVERNMENT) \cite{regcn}. These high-level entities can hardly align with events that have specific meaning during the evaluation. 

\begin{table}[h]
\centering
\fontsize{8}{10}\selectfont
\begin{tabular}{lccccc}
\hline 
\textbf{Dataset} & \textbf{$|\mathcal{E}|$} & \textbf{$|\mathcal{R}|$} & \textbf{$|\mathcal{T}|$} & \textbf{$|\mathcal{D}|$} & $Timestamp$ \\
\hline
ICEWS14	     &7,128      &230      &365       &90,730       &$point$      \\ 
ICEWS05-15   &10,488     &251      &4,017     &479,329      &$point$      \\ 
ICEWS18	     &23,033     &256      &304       &468,558      &$point$      \\ 
GDELT	     &500        &20       &366       &3,419,607    &$point$      \\ 
YAGO15k      &15,403     &32       &169       &138,048      &$interval$   \\ 
WIKIDATA     &11,153     &96       &328       &150,079      &$interval$   \\
\hline\end{tabular}
\caption{Statistics of TKGC benchmark datasets.}
\label{tab:dt}
\end{table}

\subsubsection{Evaluation Protocol}
The evaluation essentially measures how accurate a TKGC method can predict, and is normally defined as the ranking of the factual score of the true prediction among all candidates. The widely used accuracy metrics include $Hits@k$, $Mean\ Ranking$ (MR) and $Mean\ Reciprocal\ Ranking$ (MRR).
Entity prediction is basic link prediction for TKGC and is expressed as two queries, $(?,r,t,\tau)$ and $(h,r,?,\tau)$, which are predicted as $\argmin_{\hat{h}\in \mathcal{E}}q(\hat{h},r,t,\tau)$ and $\argmin_{\hat{t}\in \mathcal{E}}q(h,r,\hat{t},\tau)$, respectively. 

For some TKGC methods \cite{knowevolve} that adopt irreversible reasoning processes (e.g., path-finding from the head entity to the tail entity \cite{xerte}), the query $(?,r,t,\tau)$ instead is predicted as $(t,r^{-},?,\tau)$ by introducing $r^{-}$ (the augmented inverse relation of $r$). During training, $r^{-}$ is learned by generating an inverse fact $\{t,r^{-},h,\tau\}$ for each fact $\{h,r,t,\tau\}$ in the training set. In addition to entity prediction, relation prediction ($\{h,?,t,\tau\}$ \cite{regcn}) and time prediction ($\{h,r,t,?\}$ \cite{ttranse}) are further included in some existing works, with same metrics for the accuracy measurement.
The current research community mainly focuses on evaluating TKGC models using queries that have timestamps seen in the dataset; while a rising challenge is to further evaluate with unseen timestamps. Unseen timestamps can be categorized into two types, i.e., future timestamps \cite{regcn} and missing timestamps \cite{diac}. Future timestamps aim at measuring the effectiveness of out-of-sample prediction, and missing timestamps focus more on knowledge imputation.

\section{Temporal Knowledge Graph Completion Methods}
To predict missing links, many KGC methods learn low-dimensional representations for entities and relations of knowledge graphs, by adopting factual score functions to measure the correctness of facts and negative samples. For example, the score function of TransE \cite{transe}, $q(h,r,t)=\|e_{h}+e_{r}-e_{t}\|$, regards $r$ as the translation between $h$ and $t$. For TKGC, facts become quadruples due to additional timestamps. Therefore, TKGC expects more flexible models to further learn the temporal dynamics of knowledge graphs with the provided timestamps of facts. We observe that many TKGC methods are built upon existing KGC methods, and the main challenge is to design effective strategies to incorporate the timestamps into the factual score functions. Therefore, we analyze existing TKGC methods based on different strategies for timestamp integration.

\subsection{Timestamp-included Tensor Decomposition}
Tensor decomposition has been effective for KGC methods because it is light weighted and easy to train \cite{acmsurvey}. For these methods, a knowledge graph can be viewed as a 3-dimensional adjacent matrix; namely, a 3-way binary tensor. The three tensor modes represent the indices of head entity, relation, and tail entity, respectively, and an entry is 1 if the corresponding fact exists.
The representations of entities and relations are learned by decomposing the tensor into low-dimensional matrices. 
Considering timestamps as an additional mode of tensor (knowledge graph becomes a 4-way tensor), these tensor decomposition methods can naturally be extended for TKGC, and low-dimensional representations for timestamps are learned for the score measurement. We summarize relevant TKGC methods based on different tensor decomposition techniques used.

\subsubsection{Canonical Polyadic Decomposition}
Canonical polyadic (CP) decomposition \cite{cpd} decomposes the target tensor as the sum of a series of rank-one tensors. For a 3-way tensor $\mathcal{X}\in \mathbb{R}^{n_{1}\times n_{2}\times n_{3}}$, CP decomposes it by $\mathcal{X}\approx\sum^{d}_{\alpha=1} A_{:,\alpha}\otimes B_{:,\alpha}\otimes C_{:,\alpha}$, where $\otimes$ is the tensor outer product, $A\in \mathbb{R}^{n_{1}\times d}, B\in \mathbb{R}^{n_{2}\times d}$ and $C\in \mathbb{R}^{n_{3}\times d}$ are factor matrices. Each entry of $\mathcal{X}$ can be approximated as $\mathcal{X}_{ijk}\approx \langle a_{i},b_{j},c_{k}\rangle=\sum_{\alpha=1}^{d}a_{i\alpha}b_{j\alpha}c_{k\alpha}$. 
\cite{T-DistMult} adopts CP decomposition for TKGC by regarding a knowledge graph as $\mathcal{G}\in \mathbb{R}^{|\mathcal{E}|\times |\mathcal{R}|\times |\mathcal{E}|\times |\mathcal{T}|}$. Then, the factual score function becomes $q(s)=\langle e_{h},e_{r},e_{t},e_{\tau}\rangle$. An $imaginary$ timestamp is adopted for static facts so that the model is also capable of learning from facts without timestamps. A similar CP decomposition model is adopted by \cite{TNTComplEx}, except that it uses complex-valued representation vectors so that the model can adapt to asymmetric relations. These two models further adopt temporal smoothness penalties to ensure neighboring timestamps learn similar representations. To further improve the expressiveness of representations, \cite{telm} moves beyond complex-valued representations and learns multivector representations with CP decomposition, considering that the 2-grade geometric product expresses greater interactions among entities, relations, and timestamps. The model flexibly adapts to timestamps of both points and intervals by a dual multivector relation representation, which separately represents the start and end time of a fact. A temporal smoothness penalty with respect to geometric product is developed for timestamps and is extended to a more general autoregressive manner, instead of pair-wise neighboring timestamps.

\subsubsection{Tucker Decomposition}
Tucker decomposition \cite{tuckerorg} is another tensor decomposition technique that is recently introduced for TKGC. Typically, Tucker decomposition is regarded as a more generalized tensor decomposition technique and CP decomposition is its special case. For Tucker decomposition, a tensor is factorized into a core tensor multiplied by a matrix along each mode, i.e., $\mathcal{X}\approx \mathcal{W} \times_{1} A \times_{2} B\times_{3} C$, where $A\in \mathbb{R}^{n_{1}\times d_{1}}$, $B\in \mathbb{R}^{n_{2}\times d_{2}}$ and $C\in \mathbb{R}^{n_{3}\times d_{3}}$. A entry is approximated as $\mathcal{X}_{ijk}=\sum_{\alpha=1}^{d_{1}}\sum_{\beta=1}^{d_{2}}\sum_{\gamma=1}^{d_{3}}\mathcal{W}_{\alpha\beta\gamma}a_{i\alpha}b_{j\beta}c_{k\gamma}$. $\mathcal{W}$ is the core tensor that represents the level of interactions among decomposed matrices. Tucker decomposition is equivalent to CP decomposition when $\mathcal{W}$ is super diagonal and $d_{1}=d_{2}=d_{3}$. 
Similar to CP decomposition TKGC methods, \cite{tucker} adopts the Tucker decomposition by regarding knowledge graph as 4-way tensor. Given a fact $\{h,r,t,\tau\}$, it scores its correctness as $q(s)=\langle \mathcal{W}; e_{h},e_{r},e_{t},e_{\tau} \rangle$, where $\mathcal{W}\in \mathbb{R}^{d_{e}\times d_{r}\times d_{e}\times d_{\tau}}$ measures the interactions among entities, relations and timestamps. $\mathcal{W}$ also is interpreted as a high-dimensional linear classifier that can distinguish facts from negative samples. By adopting Tucker decomposition, the flexibility of representations is improved, because the limitation that the embedding dimensions of entity, relation and timestamp must be the same is relaxed.

\subsection{Timestamp-based Transformation}
Typical KGC methods learn static entities and relations representations for link prediction. However, this is no longer appropriate when timestamps of facts are available in knowledge graphs, because the contexts of entities and relations in the knowledge graph are different at different times. To address this issue, many TKGC methods regard timestamps as a transformation to learn entity and/or relation representations corresponding to the specific time.

\subsubsection{Synthetic Time-dependent Relation}
Facts are quadruples in TKGC, with timestamps indicating when the relation between the head entity and tail entity holds, e.g., $\{Lakers, championOf, NBA, 2010\}$. That is different from KGC that adopts facts as triples; however, if manage to convert quadruples into triples, existing KGC models can be conveniently applied. A straightforward way is to create synthetic time-dependent relations by concatenating relations with timestamps (e.g., $championOf$:$2010$). With the synthetic relations, the example fact now becomes a triple $\{Lakers, championOf$:$2010, NBA\}$.

Synthetic relations are initially adopted by \cite{ttranse} for the factual scoring measurement, i.e., $q(s)=\|e_{h}+u(r,\tau)-e_{t}\|$, where $u(r,\tau)$ is the fusion function. It evaluates the predictive performance of three basic fusion functions of $u(r,\tau)$, i.e., $e_{r:\tau}$, $e_{r}+e_{\tau}$ and $p_{\tau} e_{r}$ ($p_{\tau}\in (0,1]$ is a learnable coefficient), and the experimental results show that $u(r,\tau)=e_{r}+e_{\tau}$ is the most effective. Different from time points, time intervals of different facts may overlap, e.g., $2010$-$2014$ and $2012$-$2016$. Therefore, directly concatenating time intervals with relations may lead to more synthetic relations than necessary. To address this, \cite{splitme} discovers optimal timestamps to concatenate relations by $Splitting$ or $Merging$ the existing time intervals. For a relation, $Splitting$ splits time intervals by adopting change-point-detection (CPD) on the time series \cite{changepoint} that represent the evolution of relation; while $Merging$ iteratively merges time intervals (starting from the shortest time intervals) if related facts are not violated.

Rather than directly concatenating relations and timestamps, other methods create more expressive synthetic relations by further exploiting the explicit date of timestamps. 
\cite{ta-transe} represents the concatenation of relation and timestamp as a sequence of tokens, e.g., $\{championOf, 2y, 0y, 1y, 0y\}$ ($y$ means year) for $championOf$:$2010$. The representation of synthetic relation is obtained as the final hidden state of an LSTM \cite{lstm}, with the tokens as the input. Since LSTM can process vary-length sequences, a natural advantage is that the synthetic relation is adaptive to timestamps of different formats, such as points, intervals, or modifiers (e.g., "occursSince"). 
\cite{3drte} argues that different relations rely on different time resolutions. e.g., a person's life span is generally in years while the birth date should be in days. To achieve adaptive time resolution, the representation of synthetic relation is obtained by adopting multi-head self-attention on the timestamp-relation sequence. 

\subsubsection{Linear Transformation}
In a knowledge graph, the implications of entities and relations at different times may significantly change. To capture the implication at a specific time, timestamps are regarded as linear transformations that can map entities/relations to the corresponding representations. Specifically, \cite{hyte} regards timestamps as hyperplanes, which segregate the temporal space into discrete time zones. The entity/relation representation corresponding to a time zone is obtained by the projection with the hyperplane. For timestamp $\tau$, the hyperplane is defined as $w_{\tau}\in \mathbb{R}^{d}$ and $\|w_{\tau}\|=1$. The projection function of $\tau$ is defined as $\mathcal{P}(e)=e-w_{\tau}^{T}ew_{\tau}$, where $e$ is the static representation of entity or relation. The factual score is calculated with the projected representations as $q(s)=\|\mathcal{P}(e_{h})+\mathcal{P}(e_{r})-\mathcal{P}(e_{t})\|$. To improve the expressiveness for multi-relational facts, \cite{hhyte} includes an additional relational matrix to map entities to be relation-specific, before being projected by the hyperplane. Moreover, \cite{timespan} adopts GRU \cite{gru} on the sequence of hyperplanes to further capture the dynamics among hyperplanes. 

Other than hyperplanes, \cite{tero} regards the transformation for entities as a linear rotation in the complex space. That is, $e_{h\tau}=e_{h}\circ e_{\tau}$ and $e_{t\tau}=e_{t}\circ e_{\tau}$, where $\circ$ is the Hermitian product in the complex space. Then, the relation is regarded as the translation of the rotated head entity to the conjugate of the tail entity as $q(s)=\|e_{h\tau}+e_{r}-\bar{e}_{t\tau}\|$. To achieve arbitrary time precision, \cite{toke} first encodes a timestamp into a one-hot vector; and different regions in the vector represent different time resolutions, e.g., centuries or days. The one-hot vector of the timestamp corresponds to a series of linear transformation matrices, which are used to map entities/relations to be time-dependent.

\subsection{Dynamic Embedding}
In the TKGC task, time-dependent representations are expected to exhibit the change of the implications and background of entities and relations over time. Notably, the representations of an entity/relation along the timeline are not independent, but usually follow specific dynamical evolution patterns. For example, a person's life cycle can only be  $bornIn\xrightarrow{}workAt\xrightarrow{}dieIn$, and it is irreversible. Dynamic embedding methods aim at capturing these evolution patterns by encoding the dynamics in learned representations.

\subsubsection{Representations as Functions of Timestamp}
To represent the dynamical evolution of entities/relations, an intuitive method is to develop representations as functions of timestamps, which represent different types of dynamic patterns.
In \cite{atise}, the representations of entities and relations are regarded as time series, which can be decomposed into three components, i.e., $e=e_{static}+trend(\tau)+seasonal(\tau)+\mathcal{N}$. $e_{static}$ is an invariant component that represents the static feature of entities/relations, $trend(\tau)$ and $seasonal(\tau)$  (with $\tau$ as the input parameter) respectively represent the trend and seasonal features, and $\mathcal{N}$ is an addictive random noise.
\cite{dyernie} develops a similar model that adopts a static component and a time-varying component for a representation; but is defined in the hyperbolic space. The hyperbolic space can express more flexible geometric structures for the graph-structured data than the Euclidean space.
Accordingly, the interactions among entities and relations are defined as the product of Riemannian manifolds. The time-varying component of a representation, which represents the dynamical evolution of entities, is regarded as the movement on manifolds; namely, the velocity vector in the tangent space. 
Inspired by diachronic word embedding, \cite{diac} proposes a diachronic embedding for entities and relations. In this setup, the representation is divided into a static segment and a time-varying segment. Likewise, the static segment represents the time-invariant features, and the time-varying segment is a neuron that has timestamp as the input. The diachronic embedding is model-agnostic, and can conveniently incorporate relational domain knowledge to improve the prediction accuracy.

\subsubsection{Representations as Hidden States of RNN}
In contrast to specific dynamical patterns, recurrent neural networks (RNN) can adaptively learn the dynamical evolution of entities and relations.
\cite{knowevolve} models the occurrence of facts as a multidimensional temporal point process, which represents the complex co-evolution of multiple dynamic events. 
The model uses a conditional intensity function, which is implemented as the factual score measured with entity/relation representations, to infer the time of the next fact occurrence, given previously occurred facts. Meanwhile, the representations of the fact's head entity and tail entity at $\tau$ are learned as the output of two separate RNNs, respectively. The input of either RNNs includes the concatenation of head and tail entity presentations prior $\tau$, aiming at capturing their dynamical co-evolve patterns over time.
Similarly, \cite{temp} incorporates a structural encoder that implements multi-hop messaging passing and a temporal encoder with sequential encoders such as GRU or self-attention \cite{attention}. The structural encoder learns the structural dependencies of entities at each timestamp, and the outputs are further fed into the temporal encoder to integrate with the temporal dynamics. The hidden states that capture both information are adopted as dynamic entity representations. The model further tackles the temporal heterogeneity, i.e., sparsity and variability of entity occurrences, by data imputation (for inactive entities) and frequency-based gating.

\subsection{Learning from Knowledge Graph Snapshots}
With the timestamps, the original knowledge graph can be viewed as a series of knowledge graph snapshots/subgraphs, i.e., $\mathcal{G}=\{\mathcal{G}_{1},\mathcal{G}_{2},...,\mathcal{G}_{|\mathcal{T}|}\}$, with each subgraph only containing facts tagged with the corresponding timestamp. In this way, the knowledge graph becomes a temporal evolving subgraph that has varying relation connections. The link prediction problem is performed by inferring the multi-relational interactions among entities and relations over time. 

\subsubsection{Markov Process Models}
To learn the temporal dynamics, \cite{rtfe} regards the states of knowledge graph evolves over time following the first-order Markov process. That means, the state of a knowledge graph snapshot depends on its previous snapshot, through a probability transition matrix, i.e., $S_{\tau+1}=S_{\tau}\cdot P_{\tau}$. $S_{\tau}$ indicates the state of $\mathcal{G}_{\tau}$ and is defined as the combination of the representations of entities/relations and the learnable state parameters.
The model training is implemented in a recursive update manner, with the static embedding used as an effective initialization. In contrast to deterministic approaches, \cite{dbkge} adopts probabilistic entity representations, based on variational Bayesian inference \cite{vae}, to jointly model the entity features and the uncertainty. The representations are defined as Gaussian distributions, with learnable means and variances. The generation process in the model is defined as conditional probability $p(\mathcal{G}_{\tau}|\mathcal{E}_{\tau},\mathcal{R}_{\tau})$. First-order Markov rule is adopted for entities as $p(\mathcal{E}_{\tau}|\mathcal{E}_{<\tau})=p(\mathcal{E}_{\tau}|\mathcal{E}_{\tau-1})$. Relations at different timestamps are regarded as independent since they normally denote time-insensitive actions in datasets. Therefore, the joint probability can be simplified as $p(\mathcal{G}_{\leq \tau},\mathcal{E}_{\leq  \tau},\mathcal{R}_{\leq  \tau})=\prod_{i=1}^{\tau}p(\mathcal{G}_{i}|\mathcal{E}_{i},\mathcal{R}_{i})p(\mathcal{E}_{i}|\mathcal{E}_{i-1})p(\mathcal{R}_{i})$.
The expressive evolution patterns learned by the model are shown to be effective for TKGC in both offline and online scenarios.

\subsubsection{Autoregressive Models}
\cite{renet} models the dynamical evolution of facts with an autoregressive manner, i.e., the generation of a fact that belongs to $\mathcal{G}_{\tau}$ dependents on $\mathcal{G}_{\tau-m:\tau-1}$, where $m$ is the order of the autoregression. Besides $\mathcal{G}_{\tau-m:\tau-1}$ that carries the graph structures, the generation process further recurrently incorporates local multi-hop neighboring information of the fact, with a multi-relational graph aggregator. In addition to graph structures, \cite{regcn} employ multi-layer GCN \cite{gcn} on each graph snapshot to capture the dependencies of concurrent facts. The gate recurrent component is adopted to efficiently learn the long-term temporal patterns from historical facts and also alleviate the gradient vanishing. Moreover, the static properties of entities (e.g., type) are used as constraints to further refine the learned representations.
Different from the discrete evolution process, \cite{tango} adopts continuous-time embedding to encode both temporal and structural information of historical knowledge graph snapshots. Structural information is captured with multi-relational graph convolutional layers, and dynamical evolution is learned by neural ordinary differential equations (NODES) \cite{nodes}. Considering that many facts are not informative when they do not change across two adjacent timestamps, a graph transition layer is further included in the model to emphasize facts that dissolute or form across two knowledge graph snapshots.

\subsection{Reasoning with Historical Context}
The chronological order of facts in the knowledge graph is revealed by the availability of timestamps. That enables predicting missing links by reasoning with the historical context of the query. Normally, facts that occurred before and related to the query are regarded as their historical context. We observe existing methods use different perspectives to interpret the relevance between the query and its historical context for the link prediction.

\subsubsection{Attention-based Relevance}
Attention mechanism that attempts to selectively concentrate on a few important aspects can automatically capture the facts' relevance. 
Following this line, \cite{xerte} implements a reasoning process as the expansion of a query-dependent inference subgraph. The inference subgraph iteratively expands through sampling neighboring historical facts (share the same head entity). The expansion is directed to the query's interest according to the edge attention scores, which are calculated by message passing from historical facts with a temporal relational graph attention layer. The final inference subgraph is regarded as the interpretable reasoning path for the predicted results.
To better learn the long-term dependency of facts, \cite{tgap} develops a path-based multi-hop reasoning process by propagating attention through the edges on the knowledge graph (with attention flow \cite{flow}). Therefore, the inferred attention distribution is used as the natural interpretation of the prediction. Specifically, they argue that temporal displacement between historical facts and the query is more indicative than the exact timestamp; for example, ``2 days before" is more explicit than ``12/01/2021" to the query tagged with ``14/01/2021". Therefore, the model captures displacements at two different granularity, i.e., the sign of temporal displacement (past, present, and future) and the exact magnitude of the displacement. A two-stage GNN is adopted, with the temporal displacement included, so that the structural features of query-related entities and relations are also captured by the reasoning process.

\subsubsection{Heuristic-based Relevance}
Another perspective is to adopt external/domain knowledge, as heuristics or guidelines, during the relevance measurement of historical facts. Specifically, two types of predefined tendency scores ($Goodness$ and $Closeness$) are introduced by \cite{tpmod} to organize historical facts for the link prediction. $Goodness$ measures the hostility level of relation, e.g., $sanction$ is more hostile than $collaborate$, while $Closeness$ measures how cooperative two entities are. Then, historical facts are aggregated based on the tendency scores, so that more relevant clues are used for the prediction. A GRU is further adopted for the aggregated historical facts at each timestamp to learn the dynamic reasoning process.
\cite{cygnet} observes that history often repeats itself in the knowledge graph datasets; for example, they report that more than 80\% of events recorded from 1995 to 2019 in ICEWS repository have occurred previously. Based on this, the model consists of two modes ($Copy$ and $Generation$) for inference. The $Copy$ mode learns the probability that the query is a repeat of related historical facts. The $Generation$ mode learns the probability of all possible candidates to be the prediction, with a linear classifier. The outputs of the two modes are aggregated for the final prediction.

\section{Conclusion and Future Directions}
This paper presents the first overview of the recent advances on the emerging TKGC task. We first introduce the background of TKGC, and summarize benchmark datasets and the general accuracy metrics used for the evaluation. Then, we analyze existing TKGC methods based on how timestamps of knowledge graph facts are used to learn the temporal dynamics for the link prediction. In addition, considering the limitations of existing methods, we attempt to point out several promising directions for future TKGC research.

\subsection{Incorporate External Knowledge}
Although many methods have shown up, there is still a lot of room for improvement in the prediction accuracy, especially on GDELT dataset. Many aspects related to the datasets limit the performance; for example, the in-balanced distribution of facts, which results in a long tail structure for entities and relations \cite{oneshot}. A potential solution is to incorporate external knowledge to enrich the limited structural/temporal information during the model learning. For example, relational domain knowledge (e.g., $parentOf$ is the inversion of $childOf$) \cite{diac} enables rarely occurred relation capable of learning from other related relations; entity types help more realistic representation learning by linking the semantic background with entities \cite{regcn}. Moreover, the semantics of entity/relation such as name are largely ignored by existing methods. Utilize them through adopting pre-trained language models (e.g., Bert \cite{bert}) can enrich the limited information of knowledge graphs and bring external insights for the link prediction.

\subsection{Time-aware Negative Sampling}
Negative sampling assists effective representation learning of entities and relations by generating negative samples, contrasting to true facts in knowledge graphs. Generating discriminative negative samples is essential since failing to do so may hardly improve the model or even cause gradient vanishing \cite{negativesample}. Although negative sampling is an active research field for KGC (e.g., with generative adversarial network \cite{gan}), it is rarely explored in the TKGC scenario. Negative sampling for TKGC is presumably more challenging due to the additional time dimension, which requires properly tackling the complex interactions between facts and timestamps.

\subsection{Larger-scale Knowledge Graph}
Compared with datasets used for the evaluation of TKGC methods, real-life knowledge graphs are much larger and often contain billions of facts. Unfortunately, training TKGC models with current benchmark datasets are already quite painful (takes hours to days); that makes applying them on real-life knowledge graphs unthinkable. To improve the efficiency, distributed models that perform on multi-node computation resources without significantly undermining the link prediction accuracy need to be investigated, and that brings challenges such as optimal dataset partition and distributed computation (with timestamps considered). Meanwhile, the parameter sizes of existing methods are large since a unique embedding is learned for each entity/relation. However, considering that entities/relations share many similar features, developing compositional embedding \cite{composition} that represents entities/relations as the composition of a much smaller group of explicitly/implicitly shared features are promising to be explored.

\subsection{Evolutionary Knowledge Graph}
While most existing methods perform TKGC on an invariant dataset, real-life knowledge graphs are constantly evolving, through the deletion of wrong facts and inclusion of new facts. In this way, knowledge graphs will constantly have updated sets of entities, relations, and timestamps. To avoid training a new model from scratch for each knowledge graph update, TKGC should be regarded as an incremental or continual learning problem. Frontier work has attempted to address the $catastrophic\ forgetting$ in this streaming scenario with experience replay and knowledge distillation \cite{tie}, and achieves results comparative to baselines. In the future, other continual learning techniques such as regularization and progressive neural networks \cite{continual} can further be investigated for TKGC.

\bibliographystyle{named}
\bibliography{ref}

\end{document}